# Is Training Data Quality or Quantity More Impactful to Small Language Model Performance

Aryan Sajith[1[0009-0002-0230-6946]] and Krishna Chaitanya Rao Kathala[1[0009-0004-7708-561X]]

[1] University of Massachusetts Amherst, Amherst MA 01003, USA
{asajith, kkathala}@umass.edu

**Abstract.** This study investigates the relative impact of training data quality versus quantity on the performance of small language models (SLMs), utilizing the TinyStories dataset for empirical analysis. Analysis of dataset variations with respect to size (25% and 50% of the original size) and duplication (controlled rates of 25%, 50%, 75%, and 100%) were performed. Model performance was evaluated based on the validation loss, accuracy, and perplexity metrics. Results indicate training data quality plays a more significant role in the overall performance of SLMs, especially given scale of this experiment. Minimal duplication positively impacted model accuracy (+0.87% increase in accuracy at 25% duplication) without significantly increasing perplexity (+0.52% increase going from 0% to 25% duplication) but excessive duplication led to pronounced performance degradation (-40% drop in accuracy at 100% duplication). The implications of this exploration extend beyond just model performance; training large-scale models imposes significant financial and computational burdens, which can be prohibitive for organizations, individuals, and the public at large, especially in developing countries. Additionally, the energy consumption associated with large-scale training raises environmental concerns. Understanding the relative importance of data quality versus quantity could democratize AI technology, making advanced models more accessible and sustainable for all.

## 1 Introduction

Large Language Models (LLMs) have revolutionized the field of Natural Language Processing (NLP), demonstrating remarkable capabilities across a myriad of tasks such as problem-solving, logical reasoning, and summarization by developing a statistical distribution over a corpus of text and effectively predicting subsequent tokens [10]. However, despite their successes, the internal mechanisms of LLMs remain largely opaque, often functioning as a "black box" where their multi-equational and complex nature makes their decision-making process hard to interpret and understand [10]. This difficulty in interpretability and understandability, along with significant bottlenecks in the scale of training data available for training LLMs, has contributed to various experiments on the comparative importance of training data quantity versus training data quality for LLM performance [1-15].

Recent research highlights the importance of various aspects of data management, including the quantity and quality of training data [1]. The potential balance between these two critical elements can significantly impact the model's performance and accessibility, making it an essential area of further consideration. Some studies suggest that high-quality data can reduce the need for extensive data quantities, potentially lowering training costs and improving model performance [2, 14]. For instance, refined datasets with stringent filtering and deduplication outperform larger, less curated corpora [4]. Conversely, other research indicates that the sheer volume of training data remains crucial, particularly for ensuring comprehensive model coverage and minimizing overfitting [7]. Further exploration to clarify the relative importance of training data quality versus training data quantity will play a crucial role in the further progression of language models and our understanding of them.

The implications of this exploration extend beyond simply model performance. From a socio-economic perspective, training large-scale models imposes significant financial and computational burdens, which can be prohibitive for organizations, researchers, and individuals in developing countries. The cost of training and maintaining such models can limit access and contribute to an uneven distribution of language models. Additionally, the energy consumption associated with large-scale training raises environmental concerns [8]. Understanding the relative importance of data quality versus quantity could democratize AI technology, making advanced models more accessible and sustainable.

In line with related work, we define SLMs as compact neural sequence predictors (typically ≤ 100 M parameters) trained to model text distributions at character or token level [13, 15]. SLMs trade off capacity for efficiency, making them suitable for resource-constrained environments. Our line of inquiry leads to a crucial question: Is training data quantity or quality more important for Small Language Models (SLMs)? As highlighted earlier, this question has several implications. Reducing the dependency on extensive data and computational resources could facilitate the creation of more accessible and efficient language model training. By extension, exploring whether smaller models trained with high-quality data can match or exceed the capabilities of larger models is not merely an academic exercise but a pressing need with far-reaching consequences.

This work taps into small language model training by comparing the relative impact of data quantity versus quality for SLMs using the TinyStories dataset, which consists of more than 2 million stories [11]. We chose TinyStories because it was specifically created to test small language models on coherent and basic child-like text with a minute vocabulary to keep noise in check while still preserving grammar, syntax and other literary essentials [11]. With over two million synthetic short stories generated by GPT-3.5 and GPT-4, it also provides good variety and reliable volume appropriate for SLM training. Earlier work shows that models with fewer than 10M parameters can be trained to produce fluent English from TinyStories, which aligns with our desire for resource-limited training [11, 13]. Its small to medium size and open-source nature make experiments fully reproducible on commodity hardware like a M1 MacBook Pro.

Through the use of a clean, synthetic corpus rather than noisy web text, TinyStories lets us control for the effect of data quantity relative to data quality in our experiment [11].

This, in this work, we ask the question: 'Is training-data quality or quantity more important for small language models?' We tackle this question empirically by training a consistent nanoGPT-based architecture on controlled subsets of the TinyStories corpus, varying both total size (25%, 50%, 100%) and induced duplication rates (0%, 25%, 50%, 75%, 100%). Our key contributions are as follows:

1) **Controlled Data-Size vs. Data-Quality Study:** We systematically vary dataset size and duplication rate to isolate their separate effects on validation loss, accuracy, and perplexity.
2) **Empirical Evidence for Duplication Effects:** We show that a small amount of induced duplication (25%) can *improve* accuracy (+0.87%) with negligible perplexity degradation (+0.52%), whereas excessive duplication (100%) causes drastic performance collapse (-40% accuracy).
3) **Democratized Training Protocol:** We demonstrate that meaningful SLM experiments - including duplication studies - can be run on consumer-grade hardware (an M1 MacBook Pro), lowering the bar for AI research in resource-constrained settings.
4) **Open-Source Artifacts:** All code, data-preparation scripts, and dataset variations are publicly available under CDLA-Sharing-1.0, enabling easy reproduction and extension by the community.

# 2 Methodology

## 2.1 Data Source

Based on the work from Eldan and Li, showcasing that SLMs can exhibit emergent capabilities present in large language models with high quality training data, we utilized their TinyStories dataset released under the CDLA-Sharing-1.0 license for this experiment [9]. All dataset variations we created for this experiment are also open-sourced, free to access and released under the CDLA-Sharing-1.0 license [6].

## 2.2 Data Preparation

To measure the impact of dataset size we created two size-based variations from the original TinyStories dataset with 25% and 50% the number of stories from the baseline dataset. The specific stories chosen were done so stochastically as per our prepare.py scripts to avoid positional bias by choosing stories right at the beginning, the end, and so on.

To measure the impact of dataset quality we created several variations at each size level by duplicating an arbitrary story 25%, 50%, 75% and 100% of the time. The stories chosen to duplicate and those chosen to be left behind were, once again, stochastically chosen to minimize bias in the selection process. All stochastic dataset variations and python scripts for non-stochastic variations have been made publicly accessible [3, 6].

### 2.3 Model Source and Description

SLMs trade off capacity for efficiency, making them suitable for resource-constrained environments. The SLM we utilized this work was trained using the publicly released attention-based character-level nanoGPT model [12]. This model was released under a MIT license and was forked for this experiment [3]. This neural network was chosen for its simplicity, ease of setup and consistency across varying size and duplication levels.

This model leverages many well-established deep learning techniques: Forward propagation of running input sequences through a neural network to calculate loss, backward propagation computation of gradients to update model weights via gradient descent to minimize loss, incrementally changing learning rate due to learning rate decay, and a multi-layer architecture with attention heads and embeddings appropriate for a character-level predictive model [3]. While detailed training configurations and auxiliary settings are available in the accompanying code repository [3], key model components and parameters are summarized below for reference:

**Table 1.** Key model and training parameters

| Hyperparameter | Value |
|---|---|
| Layers | 4 |
| Embedding Size | 128 |
| Number of Parameters | 7.2 million |
| Dropout | 0% |
| Batch Size | 32 |
| Block Size | 128 |
| Learning Rate | 0.0001 |
| Maximum Number of Iterations | 1000 |
| Evaluation Interval | 50 |
| Evaluation Iterations | 100 |

### 2.4 Computer Specs and Setup

An M1 MacBook Pro with 16 gigabytes of RAM, 1 terabyte of storage, 10 core CPU, and 16 core GPU was used for training these models. The laptop was fully charged and plugged into a power source while training these models

### 2.5 Experimental Protocol

Our empirical investigation followed a systematic protocol to isolate the effects of data quantity and quality on SLM performance. The complete, reproducible workflow is available in our public code repository [3] and is summarized below:

First, we established a baseline dataset by processing the original TinyStories corpus [9]. This involved tokenizing the text and creating initial training and validation sets in a binary format optimized for model consumption.

Next, we systematically generated a suite of experimental dataset variations to analyze controlled variations in both data quantity and quality:

- **Data Quantity:** To study the impact of dataset size, we created subsets containing 25% and 50% of the stories from the baseline dataset, selected stochastically to prevent positional bias.
- **Data Quality:** To study the impact of data integrity, we introduced controlled levels of duplication (25%, 50%, 75%, and 100%) into each size-based subset by stochastically selecting and replicating a single story to achieve the target rates.

The model architecture and all training hyperparameters (as detailed in Table 1) were held constant across all experiments to ensure that any observed differences in performance could be attributed directly to the variations in the training data. Additionally, model checkpoints and evaluation metrics (validation loss, accuracy, and perplexity) were logged for further investigation.

## 3 Analysis

### 3.1 Model Evaluation

The three main measures of model evaluation used were loss, accuracy, and perplexity relative to the evaluation set. Loss acts as a measure of the model prediction inaccuracy relative to the true output, so lower loss is better. Accuracy acts as a measure of correctly predicted next characters in the sequence, so higher accuracy is better. Perplexity acts as a measure of how "surprised" the model was based on the observed true output, so lower perplexity is better. These measures were calculated within the main train.py file during the evaluation step of the procedure by outputting to a file.

### 3.2 Data Evaluation

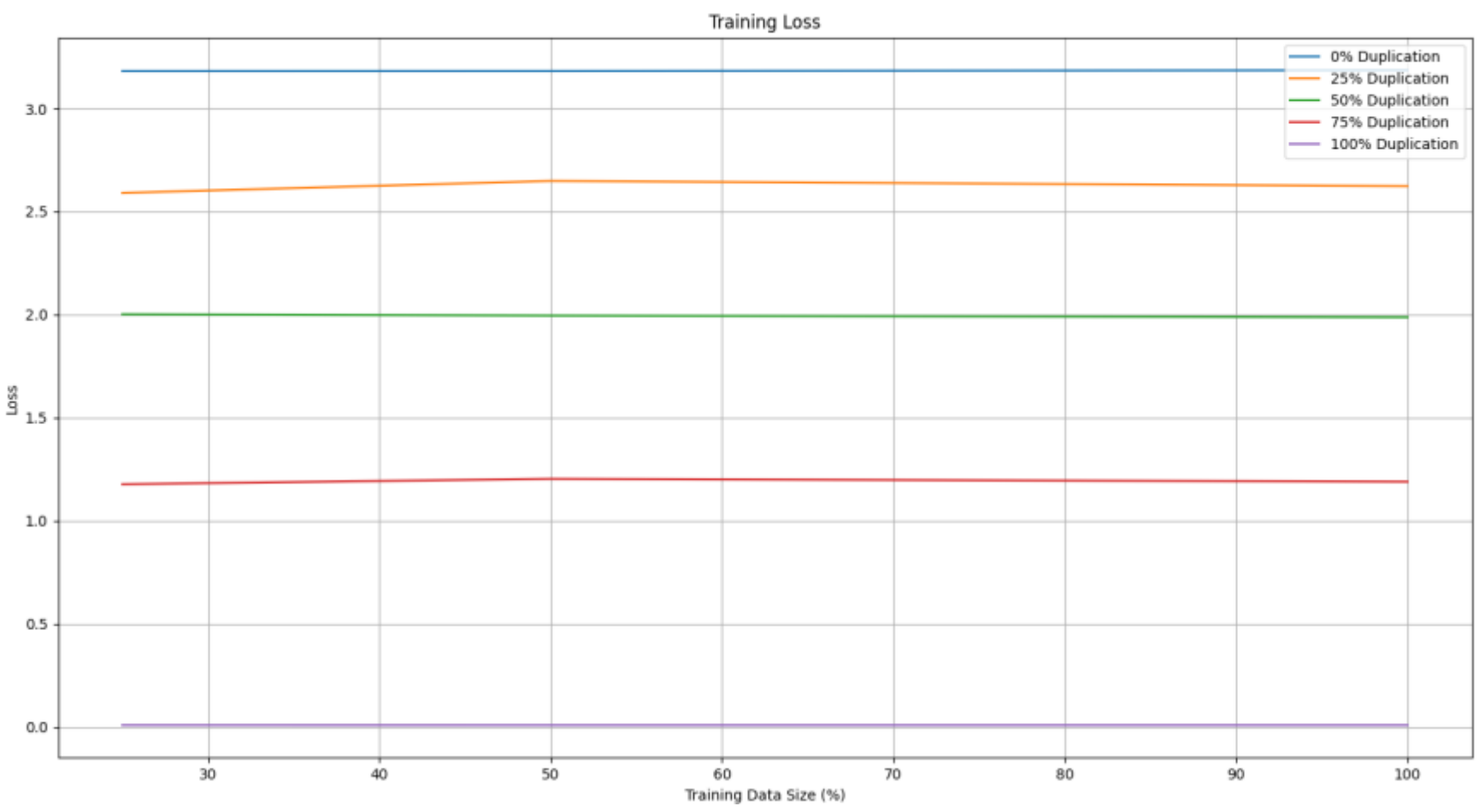

**Fig. 1.** Training Loss Data – Measured training loss across various size and duplication levels.

**Table 2.** Training loss data measured across various size and duplication levels.

| | 25% Size | 50% Size | 100% Size |
|---|---|---|---|
| 0% Duplication | 3.1821 | 3.1822 | 3.1862 |
| 25% Duplication | 2.5907 | 2.6489 | 2.6237 |
| 50% Duplication | 2.0021 | 1.9956 | 1.9877 |
| 75% Duplication | 1.1774 | 1.2035 | 1.1895 |
| 100% Duplication | 0.0084 | 0.0083 | 0.0083 |

The TinyStories dataset used for this experiment provided a high level of diversity of grammatical structures (verbs, nouns, adjectives, and so on), narrative features (dialogue, morals, plot twists, and so on), and others which is highlighted by the small degree of variance in training loss across the various size levels [11]. Training loss is analysed as per Figure 1 and Table 2. The biggest difference was noted between the 25% size and 50% size at the 25% duplication level for training loss variance of 0.0582 at about a 2.22% difference. Additionally, the improved training loss across all duplication levels indicates consistent overfitting of the model to increased duplication levels in the training dataset.

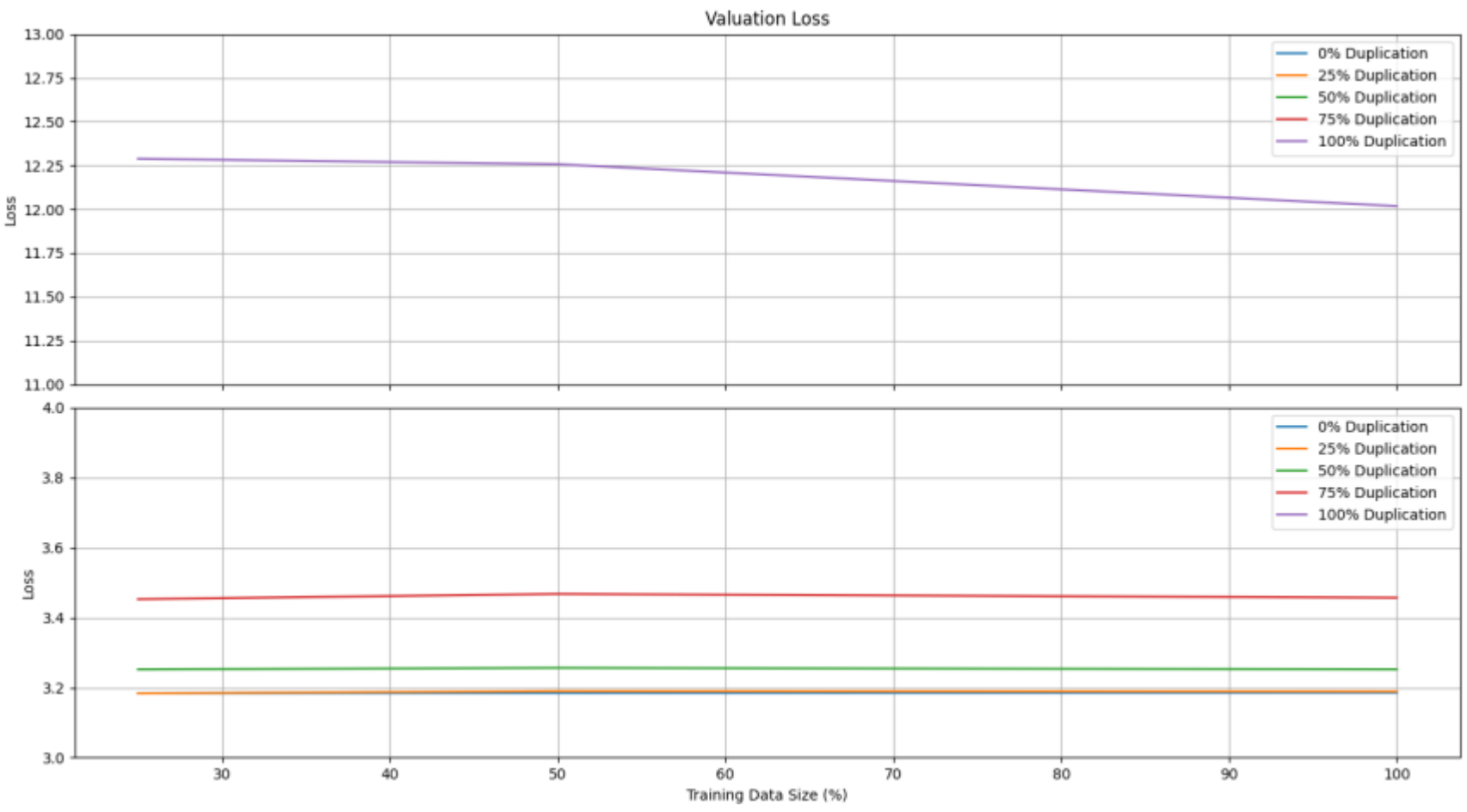

**Fig. 2.** Evaluation Loss Data - Measured evaluation loss across various size and duplication levels. A break was used on the y-axis to better visualize data trends.

**Table 3.** Evaluation loss data measured across various size and duplication levels.

| | 25% Size | 50% Size | 100% Size |
|---|---|---|---|
| 0% Duplication | 3.1838 | 3.1849 | 3.1859 |
| 25% Duplication | 3.1836 | 3.1900 | 3.1891 |
| 50% Duplication | 3.2521 | 3.2564 | 3.2522 |
| 75% Duplication | 3.4526 | 3.4647 | 3.4568 |
| 100% Duplication | 12.2886 | 12.2574 | 12.0182 |

Evaluation loss is analyzed as per Figure 2 and Table 3. The evaluation loss was largely consistent between the 0% and 25% duplication levels with the largest difference being at size 50% of 0.0051 at about a 0.16% difference. This falls in line with other work that suggests that repetition of training data negatively affects final model performance [7]. It is interesting to note, however, that evaluation loss drastically deteriorates past 50% story duplication, however, retains similar performance at the 25% duplication level and even slightly improves on the evaluation loss from 3.1838 to 3.1836 at the 25% size level. This suggests that minimal levels of duplication may not have a significant negative impact on model performance. As expected, at 100% story duplication and 25% size, the evaluation loss was significantly worse achieving 12.2886 which happened to be the maximal evaluation loss for the whole experiment and fairly similar across all size levels– indicating that with 100% duplication of stories the model overfits significantly albeit training size did seem to alleviate this slightly as the difference between 25% size and 100% size with 100% duplication was 0.2704 at about 2.22% better evaluation loss with increased size. This suggests that increased training dataset size can preserve model performance across datasets of varying quality, which also falls in line with other work [7].

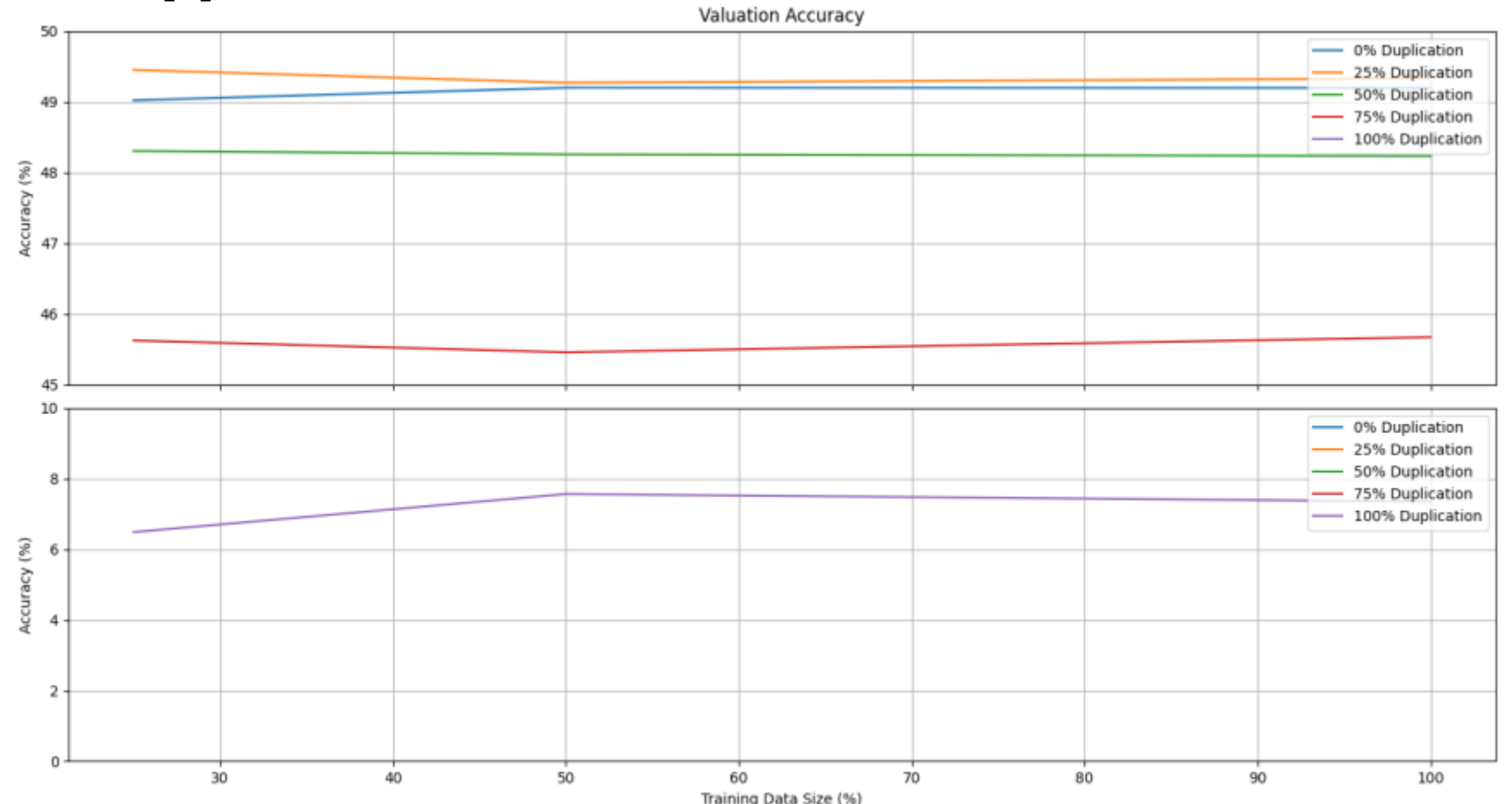

**Fig. 3.** Evaluation Accuracy Data – Measured evaluation accuracy across various size and duplication levels. A break was used on the y-axis to better visualize data trends.

**Table 4.** Evaluation accuracy data measured across various size and duplication levels.

| | 25% Size | 50% Size | 100% Size |
|---|---|---|---|
| 0% Duplication | 49.0200 | 49.1978 | 49.1947 |
| 25% Duplication | 49.4500 | 49.2684 | 49.3294 |
| 50% Duplication | 48.3034 | 48.2553 | 48.2316 |
| 75% Duplication | 45.6222 | 45.4552 | 45.6691 |
| 100% Duplication | 6.4844 | 7.5622 | 7.3472 |

Evaluation accuracy is analyzed as per Figure 3 and Table 4. Interestingly, the evaluation accuracy was improved across all sizes from the 0% duplication to the 25% duplication level. The largest increase was at the 25% size level from 49.02 to 49.45 for a 0.87% increase. This indicates that some level of story-level duplication improves the general accuracy of the model and suggests that artificially induced duplication could play a role in the training data quality for small language models. Also note that evaluation accuracy was largely consistent until the 75% duplication level where about a 3% drop in accuracy occurs. At the 100% duplication level the accuracy plummets by 40% to about 6.5%-7.5% across the various sizes. Again, this falls in line with prior results which indicate that the model overfits to the training dataset especially with limited diversity of training data due to significant levels of duplication.

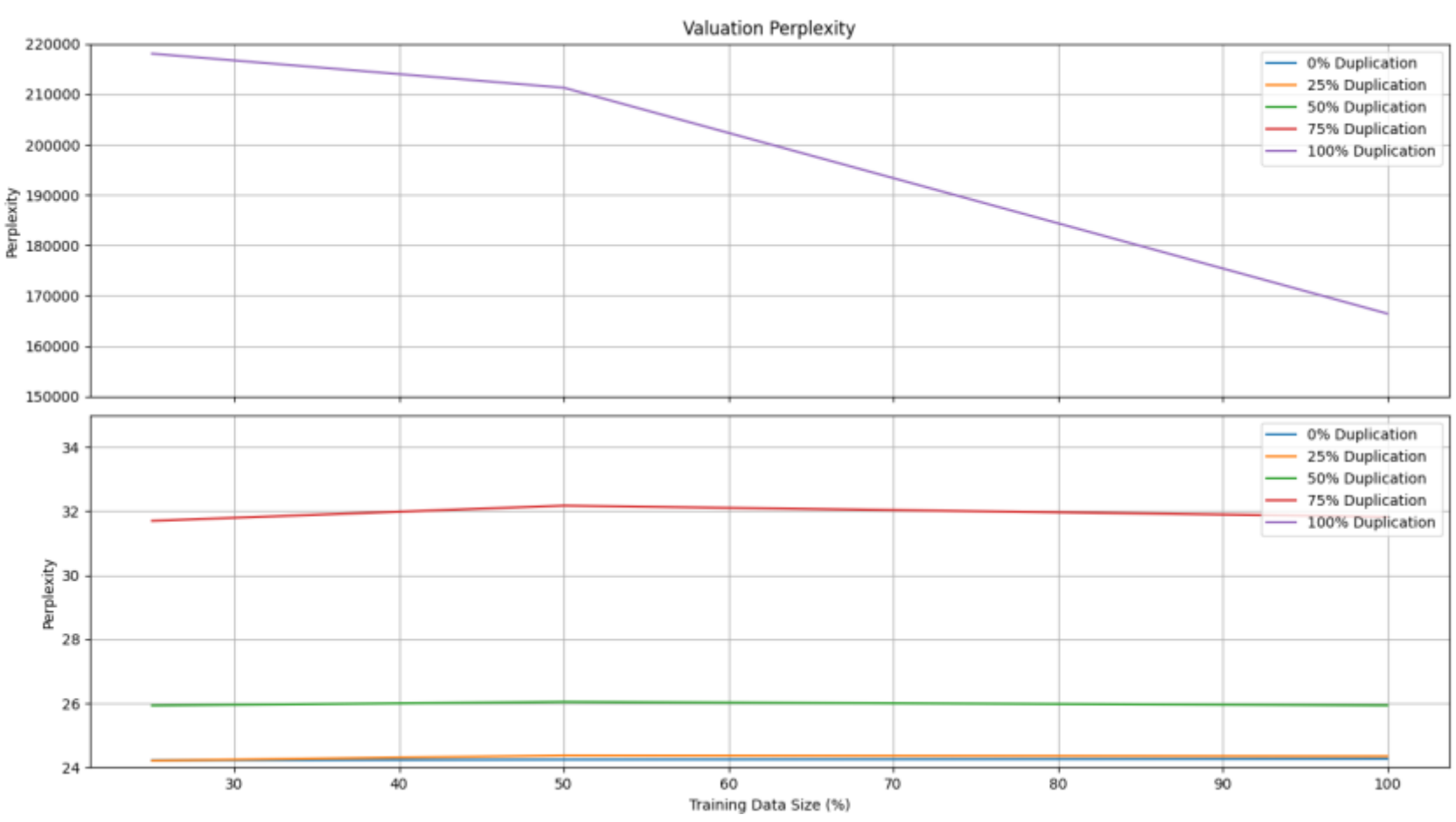

**Fig. 4.** Evaluation Perplexity Data - Measured perplexity across various size and duplication levels. A break was used on the y-axis to better visualize data trends.

**Table 5.** Evaluation perplexity data measured across various size and duplication levels.

| | 25% Size | 50% Size | 100% Size |
|---|---|---|---|
| 0% Duplication | 24.2228 | 24.2482 | 24.2707 |
| 25% Duplication | 24.2176 | 24.3762 | 24.3521 |
| 50% Duplication | 25.9363 | 26.0492 | 25.9400 |
| 75% Duplication | 31.7007 | 32.1755 | 31.8266 |
| 100% Duplication | 218,062 .2969 | 211,301.1562 | 166,426.7656 |

Lastly, evaluation perplexity is analyzed as per Figure 4 and Table 5. The evaluation perplexity was largely consistent across all sizes at 0%-25% duplication levels. This follows the prior data trends which suggest that minimal levels of duplication have limited negative effects and, in some cases, positive effects on model performance. In this case, perplexity worsened at the 50% and 100% size levels when going from 0% to 25% duplication; however, it improved at the 25% size level by 0.02%. The largest decrease across 0% and 25% duplication levels was at 50% size by 0.53%. This suggests that some minimal level of duplication has a net negligible effect on model performance and falls in line with other work [14]. As expected, a drastic falloff in perplexity occurs at the 100% duplication level whereas relatively similar levels of perplexity can be observed from 0% - 50% levels of duplication, suggesting that SLMs are either resilient to certain levels of duplication or somehow benefit from some level of story-level duplication regarding general performance.

### 3.3 Comparison with State-of-the-Art Methods

Our minimal induced duplication results (+0.87% accuracy, +0.52% perplexity on ~2 M TinyStories tokens) mirrors the perplexity-pruning gains of Marion et al. (2.1% and 1.6% accuracy improvements on GLUE/SNI with 30% of GLUE data), though it should be mentioned that their work measures performance on higher-level language understanding tasks(like sentiment classification), while ours is character-level next-token prediction in a nanoGPT setting [14]. In the same vein, SemDeDup demonstrates that deletion of 50% of LAION's semantically redundant instances reduces training time by half with minimal performance loss and enhances out-of-distribution generalization on image-captioning benchmarks, highlighting the strength of quality-based deduplication [8]. The MiniPile Challenge proved that a 6 GB filtered subset (~0.18% of The Pile's 3 300 GB) brings only a 1.9%/2.5% reduction on GLUE/SNI benchmarks which demonstrates the capability of small, high-quality corpora to match huge unfiltered datasets [2]. At web scale, Penedo et al. trained Falcon LLMs on 600 B tokens of cleaned CommonCrawl data compared to the hundreds of billions ingested by GPT-3 to match GPT-3 zero-shot performance and beat Pile-based models by up to 3.5% [4]. On code generation, Gunasekar et al.'s phi-1 model (trained on 7 B "textbook-quality" tokens) achieved 50.6% pass@1 on HumanEval with a mere 1.3 B parameters, demonstrating the skewed impact of carefully curated text [5]. Conversely, Xue et al. showed that pretraining data replication induces extreme overfitting and multi-epoch loss, which was mirrored by our 40% accuracy decline at 100% duplication [7]. These results are summarized in the Data Management survey that highlights the impact of high-quality data filtering which our findings empirically validate [1]. Zhao et al. also note further that robustness and explainability are reliant on good-curated corpora, yet again highlighting the significance of data quality [10]. Finally, the TinyStories benchmark itself proved that sub-10 M parameter models can generate fluent English, which justified our use of a 7.2 M-parameter SLM model to contrast the impact of training data quality as opposed to quantity [11].

## 4 Limitations and Future Directions

### 4.1 Model and Dataset Constraints

Firstly, experiments conducted in this study were limited to small language models (SLMs), the TinyStories dataset and a relatively low quantity of training data. For example, the largest difference between dataset size levels was observed for valuation perplexity going from the 25% size level to the 100% size level with a 26.86% improvement in model performance (can be observed in figure 4). Aside from this, no significant changes were found across varying size levels. Thus, the results may not generalize to larger language models or datasets of larger scale with varying characteristics. This limitation restricts the broader applicability of the findings to other contexts within the field of Natural Language Processing (NLP).

Secondly, the dataset variations created by stochastic selection and duplication processes were designed to avoid positional biases; however, the proposed methods may still introduce unintended biases or fail to capture the full diversity of the original dataset. Additionally, the specific variations chosen do not represent all possible forms of data quality and quantity combinations, such as toxicity filtering, data age, and others [1].

**Future Directions:**
Future work should extend these experiments to larger model architectures and more diverse, real-world datasets -while still using affordable hardware -to verify whether the quality–quantity trade-offs we observe for SLMs hold at scale. It would also be valuable to investigate other data-quality dimensions (e.g. noise filtering, domain diversity, temporal splits) and compare their effects directly against raw dataset volume.

### 4.2 Experimental Setup

Firstly, the training was conducted on a laptop, specifically an M1 MacBook Pro, with 16 GB of RAM, a 1 TB SSD, a 10-core CPU, and a 16-core GPU. While this setup offers more computational power than traditional lower-end hardware, which opens even more interesting and excellent opportunities for alternative or lower-end hardware extensions to our work, it still does not match the specialized, high-end GPUs typically used in deep learning research. These hardware limitations likely impacted both the efficiency of the training process and the final model performance. However, the successful use of an M1 MacBook Pro highlights an important consideration: Meaningful progress in language model development can be made without access to prohibitively expensive, dedicated GPUs.

This finding is particularly significant in the context of accessibility. By demonstrating that lower-end computers, possibly equipped with alternative CPUs and GPUs, can still be effective for training models, we broaden the potential for participation in AI research. Individuals or institutions with limited resources can contribute to the field, promoting a more inclusive and diverse research community. Expanding the scope and access of language model usage has even wider social implications. As language models become more integral to various applications, from education to healthcare, ensuring that these technologies are available to a wider population is crucial. The ability to develop and deploy models in local languages and cultural contexts can significantly enhance the relevance and effectiveness of AI in addressing community specific challenges. For instance, in regions where linguistic diversity is high, the ability to train models on localized datasets using affordable hardware can empower communities to create tools that reflect their unique linguistic and cultural needs.

**Future Directions:**

Future work should benchmark a wider range of affordable hardware - from low-end CPUs and mobile-class GPUs to edge devices and cloud instances - to map out the efficiency vs. performance trade-offs in SLM training. It would also be valuable to experiment with alternative model architectures and hyperparameter schedules on these platforms, as well as to train on localized or multilingual TinyStories variants to assess real-world applicability. Finally, incorporating downstream task evaluations and robustness metrics will help translate these hardware-constrained gains into practical AI deployments

## 4.3 Performance Metrics

While validation loss, accuracy, and perplexity provide a solid baseline for gauging SLM behavior, they omit important facets such as robustness to noise or adversarial inputs, interpretability of learned representations, and effectiveness on real-world downstream tasks. Likewise, our duplication-focused protocol does not address other common data-quality issues -such as labeling errors, domain shifts, or inherent bias - nor does it establish whether induced duplication and deduplication produce symmetric effects. As a result, these metrics and methods capture only a subset of model strengths and weaknesses.

**Future Directions:**
Future work can augment conventional testing with noise tests (e.g., noise or adversarial perturbations), interpretability testing, and task-specific performance to provide a more complete picture of the abilities of SLMs. At the same time, experimentation with diverse data-quality interventions such as noise removal, bias mitigation, or temporal splitting and explicit comparison of the impact to unprocessed data size can enlighten specific quality dimensions with the largest impact on SLM performance.

## 4.4 Long-Term Implications

While we've found helpful short-term results of training data quantity versus quality for SLM training, long-term impact analysis requires constant data curation, model retraining, and updates. Additionally for low- and middle-income communities, the capacity to sustain and keep models up to date over time can break or become unfeasible due to growing model demands.

**Future Directions:**
Future work can longitudinally track models trained under varying data-quality/quantity regimes via measuring upkeep effort, retraining frequency, and performance decay in live deployments across diverse domains. It would also be valuable to integrate downstream benchmarks (e.g., question answering, summarization) and robustness tests (e.g., adversarial or out-of-domain inputs) to assess real-world efficacy. Finally, studying cost-benefit trade-offs for different maintenance strategies (e.g., incremental fine-tuning vs. full retraining) will guide sustainable SLM adoption in low-resource settings.

## 5 Conclusion

In summary, while this study provides valuable insights into the relative importance of training data quality versus quantity for Small Language Models (SLMs), it is constrained by several factors, including model and dataset limitations, experimental setup, and evaluation metrics. Addressing these limitations in future research could enhance the understanding and application of these findings to a broader range of models and real-world applications in the accessible computing sphere.

Empirical analysis, using the TinyStories dataset, indicates that training data quality plays a more important role in the performance of SLMs than training data quantity, especially at the scale of this experiment. Minimal duplication positively impacted model accuracy (+0.87% increase in accuracy at 25% duplication) without significantly increasing perplexity (+0.52% increase going from 0% to 25% duplication) but excessive duplication led to pronounced performance degradation (-40% drop in accuracy at 100% duplication). These insights are pivotal, as they suggest that strategic improvements in data quality can offset the need for extensive data quantities, thus lowering the financial and computational burdens associated with large-scale model training. This reduction in cost and compute resources is crucial for democratizing access to language models, making them more sustainable and accessible, especially in resource constrained settings and lower-income, middle-income, and underserved communities. Further research should expand the exploration of these dynamics, considering different model architectures, broader metrics, beneficial yet accessible scale, and more to understand the full implications that training data form can have on model performance and accessibility.

**Acknowledgements.** This work was supported by the Undergraduate Research Volunteers (URV) program, offered by the University of Massachusetts Amherst, and my URV research mentor PhD Krishna Chaitanya Rao Kathala.

**Disclosure of Interests.** The authors have no competing interests to declare that are relevant to the content of this article.

## References

1. Miao, X., Jia, Z., Cui, B.: Demystifying Data Management for Large Language Models. In: SIGMOD '24: Companion of the 2024 International Conference on Management of Data, pp. 547-555. Association for Computing Machinery, New York, NY (2024).
2. Kaddour, J.: The MiniPile challenge for data-efficient language models. Preprint at http://arxiv.org/abs/2304.08442 (2023).

3. URV Data Quantity vs Data Quality - TinyStories code repository, https://github.com/Aryan-Sajith/URV-Data_Quantity_VS_Data_Quality-Research/tree/main, last accessed 2025/06/05.
4. Penedo, G., Malartic, Q., Hesslow, D., Cojocaru, R., Alobeidli, H., Cappelli, A., Pannier, B., Almazrouei, E., Launay, J.: The RefinedWeb Dataset for Falcon LLM: Outperforming Curated Corpora with Web Data Only. In: A. Oh, T. Naumann, A. Globerson, K. Saenko, M. Hardt, and S. Levine (eds.), Advances in Neural Information Processing Systems 36 (NeurIPS 2023) - Datasets and Benchmarks Track. Curran Associates, Inc., Red Hook, NY (2023).
5. Gunasekar, S., Zhang, Y., Aneja, J., Mendes, C.C.T., Del Giorno, A., Gopi, S., Javaheripi, M., Kauffmann, P., de Rosa, G., Saarikivi, O., Salim, A., Shah, S., Behl, H.S., Wang, X., Bubeck, S., Eldan, R., Kalai, A.T., Lee, Y.T., Li, Y.: Textbooks are all you need. Preprint at http://arxiv.org/abs/2306.11644 (2023).
6. TinyStories processed datasets (Google Drive), https://drive.google.com/drive/u/1/folders/1gJi6v5nH314OkCwN8xj4oaCpGfy95GvS, last accessed 2025/06/05.
7. Xue, F., Fu, Y., Zhou, W., Zheng, Z., You, Y.: To repeat or not to repeat: insights from scaling LLM under token-crisis. In: NIPS '23: Proceedings of the 37th International Conference on Neural Information Processing Systems, pp. 59304-59322. Curran Associates, Inc., Red Hook, NY (2023).
8. Abbas, A., Tirumala, K., Simig, D., Ganguli, S., Morcos, A.S.: SemDeDup: Data-efficient learning at web-scale through semantic deduplication. Preprint at http://arxiv.org/abs/2303.09540 (2023).
9. TinyStories dataset, https://huggingface.co/datasets/roneneldan/TinyStories, last accessed 2025/06/05.
10. Zhao, H., Chen, H., Yang, F., Liu, N., Deng, H., Cai, H., Wang, S., Yin, D., Du, M.: Explainability for large language models: A survey. ACM Transactions on Intelligent Systems and Technology 15(2), 1-38 (2024).
11. Eldan, R., Li, Y.: TinyStories: How small can language models be and still speak coherent English? Preprint at http://arxiv.org/abs/2305.07759 (2023).
12. NanoGPT code repository, https://github.com/karpathy/nanoGPT, last accessed 2025/06/05.
13. Lan, Z., Chen, M., Goodman, S., Gimpel, K., Sharma, P., Soricut, R.: ALBERT: A Lite BERT for Self-supervised Learning of Language Representations. In: International Conference on Learning Representations (2020).
14. Marion, M., Üstün, A., Pozzobon, L., Wang, A., Fadaee, M., Hooker, S.: When less is more: Investigating data pruning for pretraining LLMs at scale. Preprint at http://arxiv.org/abs/2309.04564 (2023).
15. Jiao, X., Yin, Y., Shang, L., Jiang, X., Chen, X., Li, L., Wang, F., Liu, Q.: TinyBERT: Distilling BERT for Natural Language Understanding. In: Findings of the Association for Computational Linguistics: EMNLP 2020, pp. 4153-4164. Association for Computational Linguistics, Stroudsburg, PA (2020).